\documentclass[letterpaper, 10 pt, conference]{ieeeconf}  

\IEEEoverridecommandlockouts                              

\usepackage{subcaption}
\usepackage{amssymb}
\usepackage{amsmath}
\usepackage{graphicx}
\usepackage{tikz}
\usepackage{svg}
\usepackage{array}
\usepackage[capitalise,nameinlink]{cleveref}

\usepackage{cite}

\title{\LARGE \bf
Team Northeastern's Approach to ANA XPRIZE Avatar Final Testing: A Holistic Approach to Telepresence and Lessons Learned
}

\author{Rui Luo, Chunpeng Wang, Colin Keil, David Nguyen, Henry Mayne, Stephen Alt, \\ Eric Schwarm, Evelyn Mendoza,  Ta\c{s}k{\i}n~Pad{\i}r, John Peter Whitney 
\thanks{All authors are from Institute for Experiential Robotics, Northeastern University, Boston, Massachusetts}
}

\begin{document}

\maketitle
\thispagestyle{empty}
\pagestyle{empty}

\begin{abstract}
This paper reports on Team Northeastern's Avatar system for telepresence, and our holistic approach to meet the ANA Avatar XPRIZE Final testing task requirements. The system features a dual-arm configuration with hydraulically actuated glove-gripper pair for haptic force feedback. Our proposed Avatar system was evaluated in the ANA Avatar XPRIZE Finals and completed all 10 tasks, scored 14.5 points out of 15.0, and received the 3rd Place Award. We provide the details of improvements over our first generation Avatar, covering manipulation, perception, locomotion, power, network, and controller design. We also extensively discuss the major lessons learned during our participation in the competition. 
\end{abstract}


\section{introduction} 

\begin{figure}[t]
    \centering
    \includegraphics[width=\columnwidth]{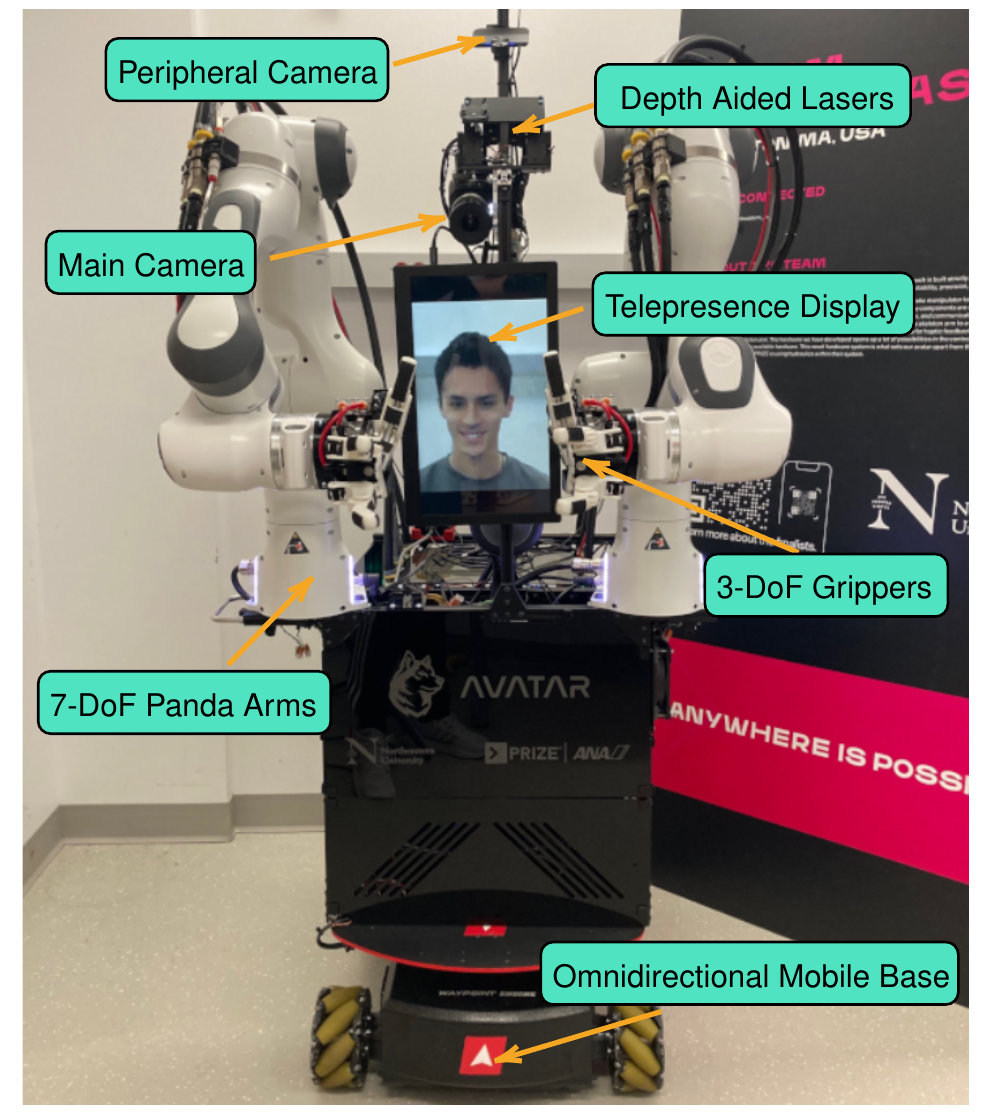}
    \caption{The Avatar robot as part of our second generation Avatar telepresence system.}
    \label{fig:default-configuration}   
    \vspace{-0.8cm}
\end{figure}

Teleoperation has been a prominent subject area in robotics for many years \cite{hokayem2006bilateral}. Compared to teleoperation, the concept of telepresence goes one step further by emphasizing the importance of operator ``immersiveness''. In particular, telepresence requires high-quality, multimodal sensory feedback and an interface that enables the operator to feel and control the remote robot as if it was an embodied avatar \cite{Minsky1980Telepresence}.
A typical telepresence system consists of two main components: a set of instruments for the operator to administer control and perceive sensory feedback; and the robot that acts as an emobodied avatar for the operator to explore the remote environment. For example, the Avatar robot developed in this work is presented in~\cref{fig:default-configuration}.

Numerous Avatar robots have been developed over the past twenty years \cite{10.1155/2013/902316, app12115557}. However, compared to the considerable progress in the field of visual and auditory display, or mobile navigation, a telepresence system that provides dexterous manipulation and haptic feedback is still lacking \cite{app12115557}.
In recent years, we have seen exciting progress in the development of telepresence systems that allow the operator to physically interact with and explore the remote environment through an Avatar robot, rather than acting as a passively perceive observer \cite{klamt2020remote, 8957294, tachi2020telesar}. The global ANA Avatar XPRIZE challenge \cite{xprize} was a four-year competition that further promoted such efforts\cite{schwarz2021nimbro, marques2022commodity, cisneros2022team}. For the first time, the robotics community had a platform to evaluate the task performance and subjective experience of telepresence technologies in manipulation, locomotion, and social interaction, all under real-world scenarios.

In November 2022, the top 20 performing semifinal teams competed at the ANA Avatar XPRIZE Finals testing event. The Finals competition presented challenges focused on richer haptic capabilities than that of the Semifinals. Moreover, all Avatar robots had to be untethered from both a network and power supply, as opposed to the tethered Semifinals setup.



Our participating team, Team Northeastern, was awarded 3rd place overall. While we could not finish all the tasks on Day 1, we were able to modify the mechanical design and network configuration for Day 2 due to our system's highly configurable properties. As a result, our system completed all the tasks on Day 2, making our team one of the three teams that had improved performance during their second day's run. Moreover, Team Northeastern was one of only four teams capable of completing all ten tasks within the time limit. Two final distinguishing traits of our system are that it was the only one to use hydraulic-actuated grippers and complete all ten tasks without the aid of a VR headset. 

The contributions of this paper are twofold: we first present our novel Avatar system and any notable improvements over its predecessor~\cite{luo2022towards} in response to the new challenges proposed for the ANA Avatar XPRIZE Finals; we then discuss in detail the important lessons learned on telepresence through Team Northeastern's participation in this competition. 
Considering many other teams' systems were equipped with more sophisticated design capabilities than ours, such as 24-DoF 5-finger robotic hands, state-of-the-art tactile sensors, or even a 3-DoF robotic head, we believe our team's success lies in the tradeoff between technical complexity and practical problem-solving capability. 
We aim to highlight the tradeoff by sharing our design choices as well as lessons learned from both the successes and failures encountered during the competition. In its entirety, this paper could be constructive to the whole telepresence research community.

\section{Results from ANA XPRIZE Avatar Final}

\begin{figure}[t]
    \centering
    \includegraphics[width=\columnwidth]{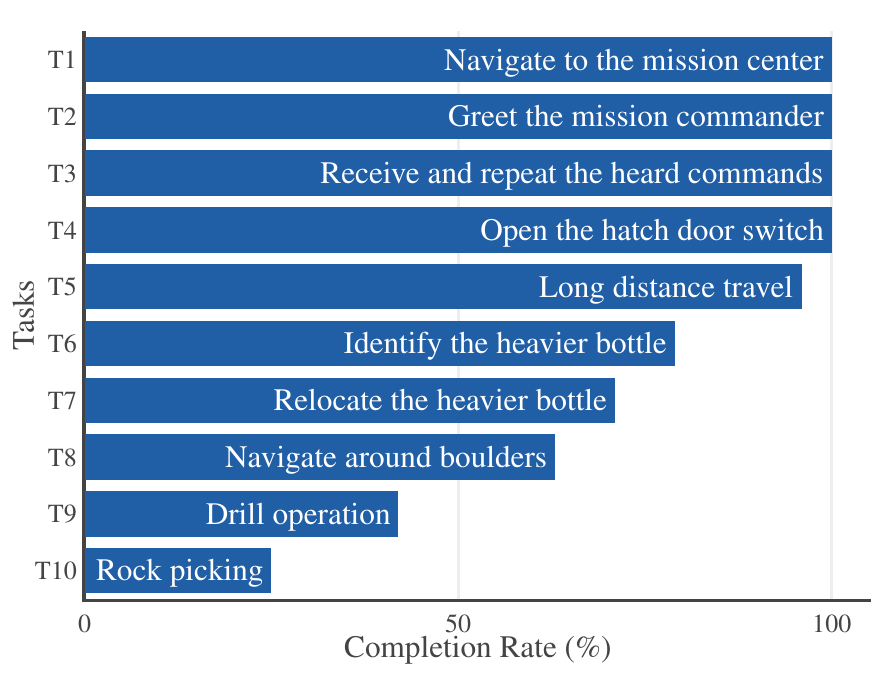}
    \caption{ANA XPRIZE Avatar Final tasks and the completion rate of 24 test runs. Task 1, 5, 8 are mobility tasks. Task 2,3 are interaction tasks. Task 4,7,9 are manipulation tasks. Task 6, 10 are haptic tasks. The statistical data is provided by XPRIZE foundation.}
    \label{fig:task-completion}
    \vspace{-0.4cm}
\end{figure}

Ten tasks were required to complete within 25 minutes sequentially, testing multiple aspects of a telepresence system's capabilities. Once the robot started the first task, no physical human intervention was allowed during the whole test run.
Each team had 45 minutes to train an operator judge who had never used their Avatar system prior.

The ten tasks can be categorized in four classes. Mobility tasks encompassed the Avatar system navigating wide, narrow, and cluttered passageways. Human-robot interaction focused on the ability of the human operator to feel present and for the human operator's presence to be perceived through the Avatar robot. Manipulation tasks required the Avatar system to flip a switch, grasp a bottle, use a drill, and pick up a small rock. 
Haptic sensing was necessary for every manipulation task, but was emphasized when the human operator was required to determine a heavier bottle from a lighter bottle and differentiate a smooth rock from a rough rock by tactile feedback only.  

The scoring system for the Final was based on a 15-point scoring system. The points were split into two parts, 10 points for completing ten well-defined tasks, and 5 subjective points given by two judges, one who controlled the Avatar robot (operator) and the other who interacted with the robot (recipient).
 
The statistical results of the task completion rate for the 12 teams that were qualified to compete in Day 1 and Day 2 final runs is summarized in \cref{fig:task-completion}. 
As seen from the results, the easiest areas of tasks were human-robot interaction and mobility. The two tasks that had the most completion rate decrease were tasks that required dexterous manipulation and haptic feedback. Only four team were able to complete the last two tasks: using the drill to unscrew a bolt, and picking the rougher rock with haptic feedback only.



\section{Avatar Generation 2 overview} 
\begin{figure}[t]
    \centering
    \vspace{0.2cm}
    \includegraphics[width=\columnwidth]{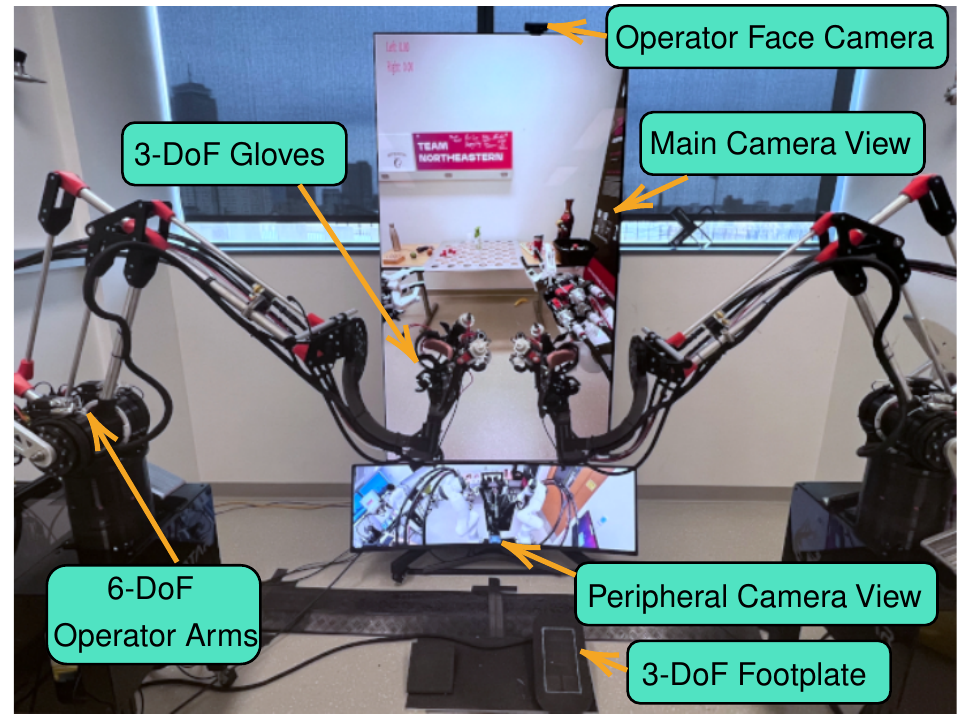}
    \caption{The new operator suite as part of our second generation Avatar telepresence system.}
    \label{fig:operator-suite}
    \vspace{-0.4cm}
\end{figure}

The second generation of our Avatar telepresence system consists of two major components: the Avatar robot (shown in \cref{fig:default-configuration}) and the operator suite (shown in \cref{fig:operator-suite}). 
As the core concepts remain the same as in our previous system, we focus on the improvements made to the new Avatar system. For a more detailed description of the previous system, readers are encouraged to refer to \cite{luo2022towards}.
Our new Avatar system has significant improvements in five aspects, which are described in the following subsections.

\subsection{Manipulation}
\begin{figure}[b]
    \vspace{-0.3cm}
    \begin{subfigure}{0.49\columnwidth}
        \includegraphics[width=\textwidth]{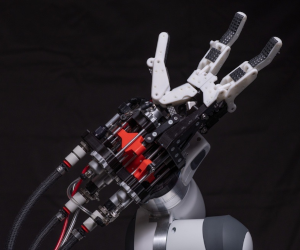}
        \caption{The 3-DoF gripper.}
        \label{fig:gripper}
    \end{subfigure}
    \begin{subfigure}{0.49\columnwidth}
        \includegraphics[width=\textwidth]{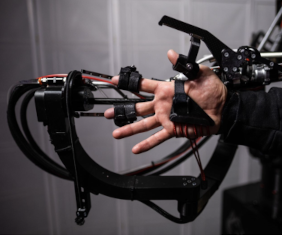}
        \caption{The operator glove.}
        \label{fig:glove}
    \end{subfigure}
    \caption{A closer look at the new 3-DoF anthropomorphic design gripper and 3-DoF operator glove.}
    \label{fig:glove-gripper}
\end{figure}

The previous manipulation system featured a single 6-DoF Universal Robotics arm as well as a 2-DoF pincer-style gripper on the robot side. The operator suite included a matching 2-DoF exoskeleton hand and a passive teleoperation arm to track position and orientation. The single gripper provided great haptic and force feedback to the operator but due to the pincer style, it had limited grips and cannot fully constrain certain objects. To improve the dexterity of the system, the grippers, and exoskeleton gloves were upgraded to a 3-DoF anthropomorphic design as shown in \cref{fig:glove-gripper}. 
Additionally, the new Avatar robot added one more gripper and arm set to feature bi-manual manipulation capabilities. These arms were switched to the 7-DoF Franka Emika Panda arms to allow for more range of motion and avoid singularity configurations. To track the operator's hand position and orientation, new operator exoskeleton arms were developed that added translational force feedback in 3 degrees of freedom. This allowed the operator to have an additional sense of interaction with the remote environment while also being able to feel the weight of objects.
\subsection{Perception} 
Our previous operator suite contained four mid-sized displays for four different viewing angles. Although the multiple viewing angles were designed to provide better coverage, we noticed that it increased the operator's mental workload and the operator preferred to focus on only one display throughout the entire task. Based on this observation, we redesigned the display setup in the operator suite as shown in \cref{fig:operator-suite}. With a vertically displaced human-size 72 inches TV as the main interaction display, the remote objects would appear to be their true sizes on the screen and the experience feels more immersive. Another advantage of positioning the display vertically is its ample vertical coverage for both bimanual object manipulation and human-to-human interaction. While the operator only needs to focus on the main display for most tasks, the bottom ultra-wide monitor provides an auxiliary view. This $180^\circ$ camera view is stitched from three cameras to provide environmental awareness for the operator.

Aural feedback is also an important channel for human sensing, therefore we setup a stereo audio system by attaching two microphones on the wrist links of two arms. It provides spatial audio and enhances the tactile sensation by amplifying the fingertip-touching sound. The auditory feedback combined with haptic feedback proved to be very helpful during the last competition task when vision is obstructed. 

Last but not least, we designed an actuated laser system to assist depth perception. By adjusting the laser angle with two servo motors, the laser line will be projected right below the center of the robotic hands and follow the hands' motion.
    
\subsection{Locomotion} 
\begin{figure}
    \includegraphics[width=\columnwidth]{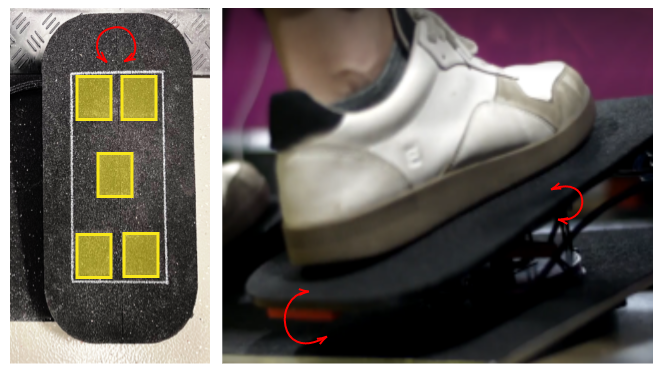}
    \caption{Top view and side view of the 3D footplate. The red arrows denote the possible control direction. The yellow boxes denote the locations of the underneath pressure sensors.}
    \label{fig:footplate}
    \vspace{-0.8cm}
\end{figure}

The mobile base of our Avatar Gen 2 system was switched from the differential drive Clearpath Husky to an omnidirectional Waypoint Vector. The capability of moving sideways greatly simplified minor pose adjustments during both locomotion and manipulation. As the operator's both arms are wearing exoskeleton arms, we designed a 3-DoF footplate for the operator to control the base with the right foot. As shown in~\cref{fig:footplate}, we installed one Vector Nav VN-100 IMU underneath the plate to measure the angular displacement and non-linearly mapped them into the twist of the omnidirectional base. 
Five pressure sensors underneath the plate detect whether the operator's foot is on the plate or not to prevent unintended robot movement caused by disturbance. 

\subsection{Power and Network} 
The requirement to make the Avatar fully untethered in the Final induced significant challenges in both power and network designs. 

From a power standpoint, our system has several key design constraints. First, we need enough battery capacity to operate untethered through the competition run while maintaining the robot's weight below 160 kg. Second, the nature of our haptics system requires a low-noise power source that can handle sudden spikes in demand (for high-frequency force feedback) without a significant voltage drop. It also requires a high enough voltage to provide peak torque while overcoming the haptic motors' back EMF. Finally, the Panda arm manufacturer does not support direct DC power solutions, so the Panda arms require an inverter.

\begin{table}[t]
\centering
\caption{Remote System Power Budget: Lists major power sinks and nominal power consumption. 
Note that estimates are slightly pessimistic to reflect variations in real world usage.}
    \label{tab:power_budget}
    \begin{tabular*}{\columnwidth}{| m{5em} | m{1.5cm} | m{4.42cm} |}
        \hline
         System & Nominal Power Draw & Notes \\
         \hline\hline
         Panda Arms & 2$\times$125W & Mean Well TS-1000 provides  AC power.\\
         \hline
         Computers & 2$\times$60-120W & We used 2 NUCs with I7-9750H CPU. No GPU is needed. \\
         \hline 
         Vector Base & 100W & Not Measured Precisely. Most of the time it is stationary.\\
         \hline
         Haptic Grippers & 20-80W & Power draw is low when not using the grippers\\
         \hline
         Other Systems & $<30$W & Includes cameras, lasers, fans, emergency stop circuitry\\
         \hline\hline
         Total & $\sim500$W & Implies an operating time of approximately 2 hours.\\
         \hline
    \end{tabular*}
    \vspace{-0.4cm}
    
\end{table}
Fortunately, the omnidirectional base Waypoint Vector turned out to have a battery system that was compatible with the above constraints. 
During untethered operation, our entire Avatar robot is powered by the Waypoint Vector base's battery system: two Valence U1-12RJ batteries connected in series, providing approximately 1kwh of power with a 29V bus at full charge. 
A rough power budget can be found in table \cref{tab:power_budget}. This allowed 2-3 hours of continuous operation depending on the specific use case. The battery system is non-removable, and the 250-watt charger requires over 5 hours to fully recharge the system. 

On the operator side, the power system is more straightforward, as most systems (computers, monitors, TV, etc.) are powered directly by generic AC sources. 
The operator haptic system has more restrictive requirements. Its motors are powered by a low-noise 48V linear supply, and sensors are powered using high-quality 18V Makita batteries.

As for the network, the Wi-Fi network presents higher latency, more jitters, and lower bandwidth when compared to the wired network. To accommodate these changes, we first switched to the UDPROS, which has much lower latency than TCPROS, for control signal transmission. Then we adopted NDI-HX \cite{ndi.tv}, a low bandwidth implementation of NDI that supports highly-compressed video formats such as HEVC, to transmit the video. The bandwidth of video transmission could be adjusted from 40 Mbit/s to 150 Mbit/s to adapt to different network conditions. The two networks were running independently on two I7-9750H NUCs on the Avatar robot with independent wireless adapters. All signals were transmitted using the on-course 5 GHz Wi-Fi between the operator and the robot.

\subsection{Controller Algorithm}

We adopted a Cartesian impedance controller for the 7-DoF Panda arm: 
\begin{equation*}
    M(q)\Ddot{q}=J^T(q)(K(x_d-x)+B(\dot{x}_d-\dot{x})-F_e),
\end{equation*}
where $q\in\mathbb{R}^7$ represents joint positions, $M(q)\in\mathbb{R}^{7\times7}$ represents inertia matrix, $J(q)\in\mathbb{R}^{6\times7}$ represents Jacobian matrix, $K\in\mathbb{R}^{6\times6}$ and $B\in\mathbb{R}^{6\times6}$ represent coupling stiffness and damping, $x_d\in\mathbb{R}^6$ and $\dot{x}_d\in\mathbb{R}^6$ represent desired position and velocity derived from communication channel, $F_e\in\mathbb{R}^6$ is the reaction force from environment. The Coriolis force $C(q,\dot(q))\dot{q}\in\mathbb{R}^7$ and gravity force $G(q)\in\mathbb{R}^7$ are compensated and not shown in the equation.
The 7-DoF Panda arm was coupled with the 6-DoF exoskeleton arm in Cartesian coordinates under the base frame such that the operator could always feel the weight of objects in the global z direction regardless of wrist angle. 
It is worth noting that the translation of the Panda arm's end-effector is coupled with the exoskeleton glove's position with bilateral force feedback, while the rotation of the Panda arm's end-effector is only following the rotation of the exoskeleton glove without force feedback.
As for the glove-gripper coupling setup, it is a similar impedance controller with bilateral force feedback coupling each DoF of grippers with the corresponding DoF of gloves.

Running along with the main impedance controller, we designed two more secondary controllers to avoid self-collision and violation of joints constraints during teleoperation: 
\begin{itemize}
    \item A nullspace controller to prevent the elbow position from drifting as well as colliding with the base was defined as 
    \begin{equation*}
        T_{null}=(I-J^T(q)J^{T+}(q))(K_{null}(q_0-q)-B_{null}\dot{q}),
    \end{equation*}
    where $T_{null}$ is the output torque, $J^{T+}$ is the pseudo-inverse of $J^T$, $K_{null}$ and $B_{null}$ are nullspace joint stiffness and damping, $q_0$ is the default joint position shown in \cref{fig:default-configuration}.
    \item A virtual wall was applied for all joints of the Panda arms to prevent joint limit errors and provide force feedback to the operator when the joint limits were nearly reached. The virtual wall was defined as
    \begin{equation*}
        \vspace{-0.1cm}
        T_{wall}=K_{wall}(q_{wall}-q)-B_{wall}\dot{q},
    \end{equation*}
    where $K_{wall}$ and $B_{wall}$ are the stiffness and damping of the virtual wall. The controller would only take effect for each joint independently when a joint exceeded its defined virtual joint limit $q_{wall}$.
\end{itemize}

Since no human intervention was allowed during the test run, we designed an error handler to  monitor software-level recoverable errors of the Panda arms, such as joint position limit violation, velocity limit violation, etc., and automatically recover the arms to the default position (shown in \cref{fig:default-configuration}) once an error occurred.

The same wave variable method~\cite{niemeyer2004telemanipulation} as in our previous system~\cite{luo2022towards} was used to encode and transmit the control signals derived above for bilateral arms and glove-grippers teleoperation. 
Although the wireless network presented new challenges for teleoperation such as a significant increment of average network delay (from $0.1$ ms to $2$ ms), more frequent frame drops, and high jitter. Adopting the same algorithm proved to be still viable because the teleoperation communication channel would remain stable as long as the average time delay remains constant over the long term. 

\begin{figure*}[t]
    \begin{subfigure}{0.32\textwidth}
        \centering
        \includegraphics[width=\columnwidth]{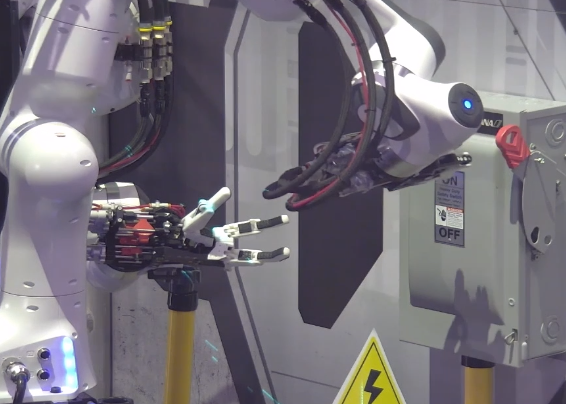}
        \caption{Flip the switch to open a door.}
        \label{fig:task-switch}
    \end{subfigure}
    \begin{subfigure}{0.32\textwidth}
        \centering
        \includegraphics[width=\columnwidth]{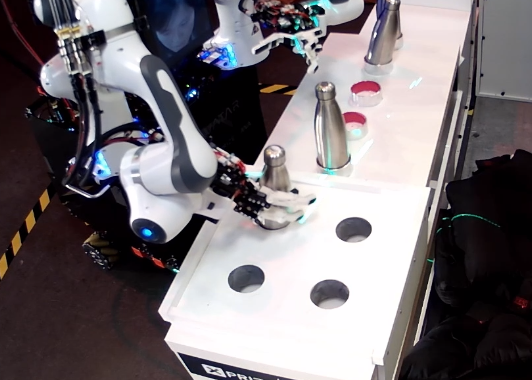}
        \caption{Identify the heavier bottle.}
        \label{fig:task-bottle}
    \end{subfigure}
    \begin{subfigure}{0.32\textwidth}
        \centering
        \includegraphics[width=\columnwidth]{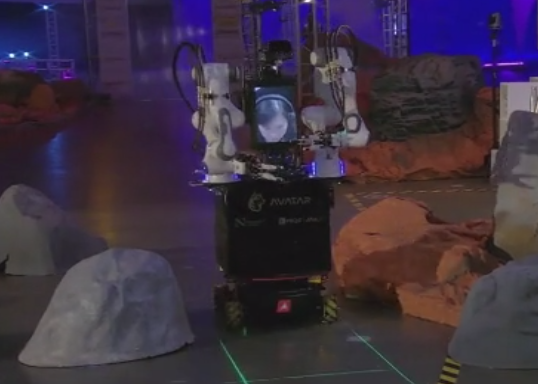}
        \caption{Navigate around boulders.}
        \label{fig:task-navi}
    \end{subfigure}
    
    \begin{subfigure}{0.32\textwidth}
        \centering
        \includegraphics[width=\columnwidth]{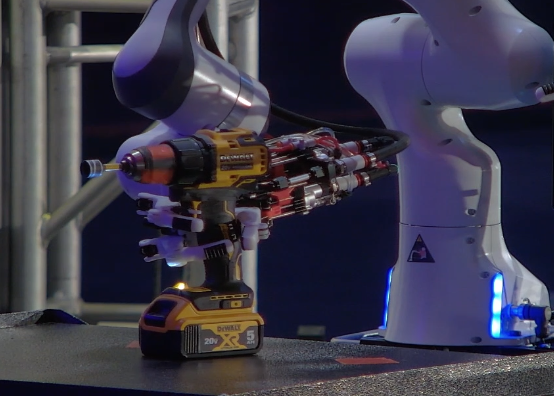}
        \caption{Pick up the drill from a table.}
        \label{fig:task-drill}
    \end{subfigure}
    \begin{subfigure}{0.32\textwidth}
        \centering
        \includegraphics[width=\columnwidth]{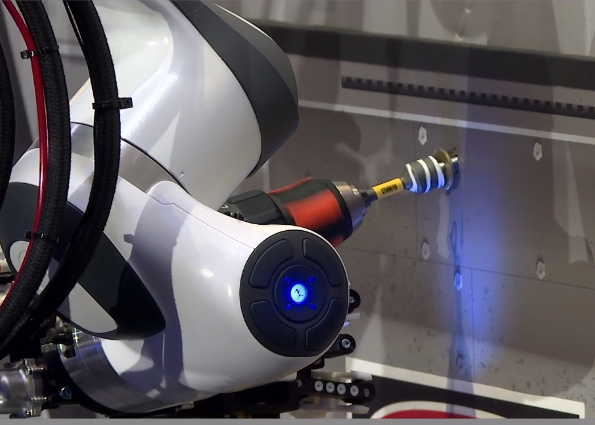}
        \caption{Use the drill to unscrew a bolt.}
        \label{fig:task-bolt}
    \end{subfigure}
    \begin{subfigure}{0.32\textwidth}
        \centering
        \includegraphics[width=\columnwidth]{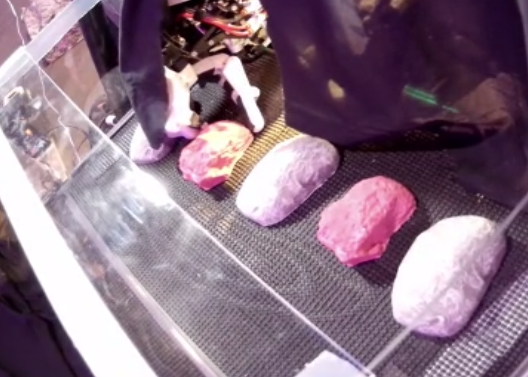}
        \caption{Identify the rougher rock without vision.}
        \label{fig:task-rock}
    \end{subfigure}
    \caption{Our Avatar system was completing the manipulation and navigation tasks in the ANA XPRIZE Avatar Final.}
    \label{fig:tasks}
    \vspace{-0.47cm}
\end{figure*}

\section{Lessons Learned from Avatar XPRIZE Final} 
In this section, we share our lessons learned from completing all the tasks in the Avatar XPRIZE Final (partially shown in~\cref{fig:tasks}), as well as in-person communication with other teams during the competition. 

\textbf{Errors are inevitable during teleoperation but ensure the system can be recovered remotely.} It is very common for robots to encounter errors during teleoperation. Torque limit violation is one of the most common errors we encountered when operating the Avatar robot. By design, a collaborative robot usually has a relatively lower joint torque or velocity limit to ensure safety during physical human-robot interaction~\cite{peshkin2001cobot}. 
Without a motion planner, it is very common for a teleoperated robot to violate the robot's constraints or collide with surrounding obstacles due to network latency, limited environmental awareness, or sudden changes in human motion. 
During the Final, our left Panda arm accidentally hit a wall during the last task and reached the torque limit of the arm, causing a safety stop that required a manual reboot to clear the safety error. Our operator judge had to continue the task without the left arm. We lost much time and 0.5 points due to the malfunction of the left arm.
A similar error occurred for Team NimbRo and Team AVATRINA because we all used Panda arms with the same safety configuration. 
However, only Team NimbRo was able to recover from that error state remotely, as they developed a comprehensive error-handling module for the arm to guarantee any error can be recovered without manual override~\cite{nimbro_com}. 
Although our own error-handling function operated well under most conditions, it was still not capable of recovering from such edge cases.

\textbf{Minimize the complexity of your control interface and information presented.}
Unlike the observation that ``operators want to control the robot at many levels'', as reported in \cite{DRC2016}, we believe that the opposite ``less is better'' viewpoint is more appropriate in designing a general-purpose telepresence system. 
The top-ranking teams followed similar design philosophies coincidentally. 
Only necessary information, such as error warnings, and object weight, is presented to the user.
The top teams' Avatar robots were controlled by the operators as if they were controlling their extended limbs, instead of by adjusting joint angles or velocities. 
Information overload could be detrimental to the telepresence experience and steepen the learning curve for a non-expert user of the system. During the Final, we only had 30 minutes to train a non-expert judge to master the complex robotic system and complete all 10 tasks in 25 minutes. 
Therefore, it is critical to rely on the existing human skillsets to teleoperate and understand the avatar, rather than providing a brand new interaction experience such as mouse clicks or joysticks.

\textbf{Impedance unmatched setup is not good for teleoperation.}
Coupling unmatched impedance (commonly different inertia) devices while maintaining one-to-one position tracking and force feedback is challenging. Equal inertia of master and slave is preferred for high transparency in bilateral teleoperation~\cite{MotoidifferentInertia}. An operator feels an oscillation of inertia on the remote Avatar robot since the virtual spring coupling between master and slave has mechanical compliance. This feeling is a disturbance compared to real contact force, and the disturbance increases if the remote avatar impedance is larger. In our avatar setup, the joint impedance of the Panda arm is more than 10 times larger than the joint impedance of the exoskeleton arm.
It was hard for the operator to differentiate the feeling between the contact force and the Pandam arm's inertia, resulting in the operator not being aware of the collision.

\textbf{Adaptable network solutions and swift deployment are important to survive an unknown environment.} 
A robust network is one of the most critical components of a telepresence system, yet it is frequently overlooked. 
In the Final, only 10 minutes were provided prior to the test run for access to the on-course network. Most teams, including us, encountered various network communication issues, such as limited network bandwidth (kilobytes per second), serious jitter, etc.. Many teams had difficulty in figuring out a solution in time due to the narrow time window for adjustments. Some teams employed complex humanoid robot systems with state-of-the-art sensors and actuators, but could not complete any of the tasks due to network challenges. Interestingly, the top teams took different approaches to network setup in the Final. Team NimbRo developed their own low-latency ROS transportation layer and utilized both 2.4 Ghz and 5 Ghz channels to dynamically deliver data packets in either channel based on current network traffic~\cite{schwarz2017nimbro}. Team Pollen and Team AVATRINA used WebRTC \cite{marques2022commodity} directly without having any network issues, which turned out to be the most compatible configuration for the on-course network. We experienced serious network jitter on Day 1 like other teams. However, because we prepared multiple alternatives for network hardware (antenna, switch, router, etc.) and various video streaming options to accommodate different network scenarios, we were able to swiftly modify our network configuration and reliably use the on-course network on Day 2. 


\textbf{Is VR the only viable option to telepresence?} 
As the only team that was able to finish all competition tasks using a non-VR system, we underwent many rounds of discussion during design on whether to use a VR headset or not. Most of the top teams used head-mounted stereo cameras to stream in real-time and render in the VR headset. Using high-efficiency video codecs like HEVC and modern GPU for encoding and decoding, the bandwidth usage is reasonably low while maintaining high-fidelity content. The motion sickness while using VR has also been greatly alleviated by disentangling the virtual camera and real-world camera motion using similar methods to those described in \cite{stotko2019vr}. VR headsets undoubtedly provide better immersiveness and depth sense in comparison to our 2D display setup. However, we note that using VR does not necessarily make teleoperation easier for all tasks. With sufficient training, the operator is able to sense depth using simply a 2D display. Moreover, our 2D display solution was appreciated by multiple judges because of its ease of usage, improved comfort, and accessibility for people who cannot use VR systems. The capability of seeing the operator's real facial expression is another advantage over synthesized facial expression, which was the only option for teams that used VR. 

\textbf{One of the biggest hurdles in telemanipulation is line of sight.}
 The most challenging and time-consuming task in the Final was to use the drill. Many Avatar robots had grippers capable of grabbing and pulling the trigger, but the operators oftentimes struggled to align the gripper fingers with the drill's trigger due to the line of sight being blocked. In most failure cases, the operator kept readjusting the mobile base of the robot to different angles, attempting to pick up the drill until trial time typically ran out. 
 We failed the task on Day 1, but as our grippers were highly customizable, we solved this issue by customizing both of our grippers with a small hook overnight before Day 2. With the adjustments, the operator would not need to align the gripper's fingers perfectly but could still pull the trigger.
 However, Team NimbRo was able to fully solve this issue by having stereo cameras installed on a 6-DoF arm, such that the operator could pan the Avatar head and look sideways. The translational motion proved to be extremely helpful in solving line-of-sight issues when compared to the 3-DoF head design in other Avatar systems.

\textbf{Sensing was multimodal, but our modality feedback was not.} 
One of the most important features of a telepresence system is its capability of providing high-quality multimodal sensory feedback \cite{Minsky1980Telepresence}. 
To complete all the tasks in the Final, participants included various sensors on the Avatar robots to obtain multimodal information, including but not limited to vision, audio, weight, and texture. 
However, when compared to the rich sensing capabilities available to the robot, the feedback side for human operators was typically lacking. 
Most teams relied on presenting all sensory readings visually, excluding audio data. 
The operators could understand the various characteristics of an object by reading sensor measurements, rather than actually perceiving them.
We believe that sensing and feedback are equally important in a telepresence system.
Regardless of the sensing capability integrated into the Avatar robot side, the operator side should have an equivalent method of experiencing this sensory source. 
By designing our own exoskeleton and glove, we were one of the few teams that could let the operator feel the actual force exerted on the arm or fingers rather than through measurement readings. 
Nevertheless, there are still many more types of sensations we could include to help improve the telepresence experience, such as temperature, fingertip tactile feedback, or even the wind force during navigation as one of the participants did. 

\textbf{How could shared autonomy benefit teleoperation?}
Incorporating shared autonomy into the teleoperation system would likely help with complex manipulation tasks. 
Surprisingly, we note that only a small portion of participating teams added autonomy to the control. 
For example, Team AVATRINA included a simple assistive feature that allowed the operator to select a virtual button in VR, initiating the autonomous grasping of the drill \cite{marques2022commodity}. 
\begin{figure}[t]
    \centering
    \vspace{0.2cm}
    \includegraphics[width=\columnwidth]{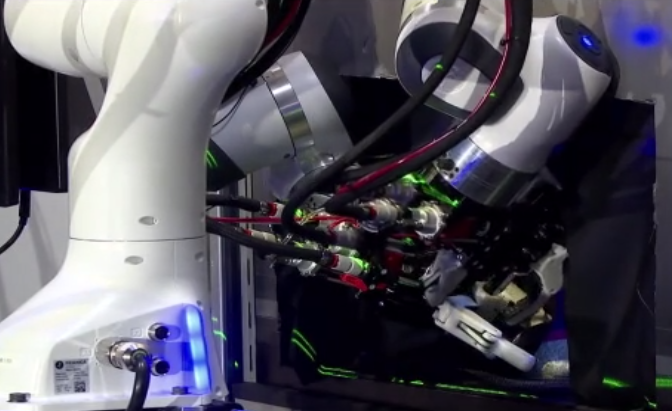}
    \caption{In the final task, the operator was required to pick up the rough rock using tactile feedback in a constrained space.}
    \label{fig:manip-constraint}
    \vspace{-0.4cm}
\end{figure}
Similar designs were applied in the control of humanoid robots where several virtual buttons for the operator to switch the humanoid robot's stance from standing during manipulation to sitting when navigating.

The lack of shared autonomy could potentially be attributed to two factors. Firstly, existing shared autonomy approaches rely on prior task knowledge, as well as multiple assumptions about human intentions \cite{losey2018review} in order to facilitate proper robotic assistance.
In a real-world scenario like the competition, the remote unknown and dynamic environment, the uncertainties of human operators, and the diversity in different tasks can all contribute to the difficulty of designing a reliable shared autonomy solution. 
Second, the advancement of dexterity in robotic hardware and control algorithms has enabled operators to perform complex manipulations directly, without requiring any assistance. 
However, we believe shared autonomy could still benefit the telepresence system in certain cases. 
For example, Team Inbiodroid's Avatar robot accidentally flipped when the operator was backing too much without knowing there were boulders behind. Other robots, including ours, encountered safety errors due to collision during manipulation in a constrained space (shown in \cref{fig:manip-constraint}). 
When the operator is focused on a task, it is hard for him or her to notice other constraints due to limited sensing capabilities. By leveraging robot autonomy for lower-level tasks, the operator could focus solely on the more challenging primary high-level task \cite{wang2015shared}.

\textbf{Unfortunately, humanoid robots still fall.} Although a bipedal humanoid robot sounds like an attractive solution for an Avatar system and many participants were using a humanoid robot as the Avatar. Achieving reliable bipedal still induced great challenges to the teams that utilized humanoid robots. 
Due to the limited view of the remote environment or unreliable network condition, teleoperated bipedal humanoids are prone to collision with surrounding obstacles under the commands of a human operator. Even worse, there is no current way of autonomously recovering the humanoid robot after control failure without human intervention. The high expense, unreliability, and lack of control software all limit the practical usage of humanoids for telepresence\cite{10035484}.  The only teams that used humanoid robots without falling were the ones that replaced legs for wheels. Still, we wish to see humanoid robots become more reliable and more dominant in the telepresence field as it provides much more dexterity when compared to wheeled robots, especially over rough terrains.

\section{Conclusions} 
In this paper, we presented the major improvements of Team Northeastern's new Avatar systems that helped award us 3rd place in ANA Avatar XPRIZE Final. Five aspects of the systems improvements were discussed in detail, including  manipulation, perception, locomotion, power/network, and controller design. We also shared multiple lessons learned throughout our participation in the competition, which could serve as a reference for future telepresence system design. By covering both aspects, we hoped to accelerate the deployment of telepresence systems in solving real-world challenges. 

\bibliographystyle{IEEEtran}
\bibliography{reference}

\end{document}